\documentclass[letterpaper, 10pt, conference]{ieeeconf}

\usepackage[left=54pt,top=50pt,right=54pt,bottom=57pt]{geometry}
\usepackage{times}
\usepackage{tabularx}
\usepackage{subcaption}
\usepackage{multirow,multicol, array}
\usepackage[font={small}]{caption}   %onehalfspacing
\usepackage{graphicx}%,dblfloatfix}
\usepackage{wrapfig}

\usepackage{algorithm}
\usepackage[noend]{algpseudocode}

\let\labelindent\relax
\usepackage{enumitem}

\usepackage{amsmath,amsthm,amssymb,amsfonts}
\usepackage{textcomp}
\usepackage{acronym}
\usepackage{balance}
\usepackage{mdwmath}
\usepackage{bm}
\usepackage{mdwtab}
\usepackage{array}
\usepackage[usenames,dvipsnames]{color}
\usepackage{eqparbox}
\usepackage{cite}

\usepackage{color}
\usepackage{psfrag}
\usepackage{epsfig}
\usepackage{url}

\usepackage{epstopdf}
\usepackage{booktabs}
\usepackage{blindtext}

\usepackage{soul}

\usepackage[colorinlistoftodos,prependcaption,textsize=small]{todonotes}

\renewcommand{\emph}{\textit}

\newtheorem*{lemma*}{Lemma}

\newtheorem*{problem*}{Problem}

\newcommand{\p}{\textbf{p}}
\newcommand{\x}{\textbf{x}}
\newcommand{\y}{\textbf{y}}
\newcommand{\z}{\textbf{z}}
\newcommand{\vv}{\textbf{v}}
\newcommand{\aaa}{\textbf{a}}
\newcommand{\e}{\textbf{e}}
\newcommand{\R}{\textbf{R}}
\newcommand{\OO}{\textbf{$\Omega$}}
\newcommand{\M}{\textbf{M}}
\newcommand{\C}{\textbf{C}}
\newcommand{\dd}{\textbf{d}}

\makeatletter
\def\maketag@@@#1{\hbox{\m@th\normalfont\normalsize#1}}
\makeatother

\IEEEoverridecommandlockouts
\graphicspath{{images/}}

\begin{document}

\acrodef{ISR}[\textsc{isr}]{Intelligence, Surveillance, and Reconnaissance}

%\title{Collision Characterization and Recovery Control for a Collision-Resilient Quadrotor Equipped with Hall Sensors}
\title{Toward Impact-resilient Quadrotor Design, Collision Characterization and Recovery Control to Sustain Flight after Collisions}

\author{Zhichao Liu and Konstantinos Karydis
\thanks{The authors are with the Dept. of Electrical and Computer Engineering, University of California, Riverside. 
	Email: \{zliu157, karydis\}@ucr.edu. %}
%\thanks{
We gratefully acknowledge the support of NSF \#  IIS-1910087, ARL \# W911NF-18-1-0266, and ONR  \# N00014-19-1-2264. %, and of the UCR Office of Research and Economic Development under a Collaborative Seed Grant.  
Any opinions, findings, and conclusions or recommendations expressed in this material are those of the authors and do not necessarily reflect the views of the funding agencies.
}}

\maketitle
\thispagestyle{empty}

%\author{\IEEEauthorblockN{Nadia Kreciglowa}
%\IEEEauthorblockA{School of Engineering and\\Applied Science\\
%University of Pennsylvania\\
%Philadelphia, PA 19104\\
%Email: nadiakre@seas.upenn.edu}
%\and
%\IEEEauthorblockN{Konstantinos Karydis}
%\IEEEauthorblockA{School of Engineering and\\Applied Science\\
%University of Pennsylvania\\
%Philadelphia, PA 19104\\
%Email: kkarydis@seas.upenn.edu}
%\and
%\IEEEauthorblockN{Vijay Kumar}
%\IEEEauthorblockA{School of Engineering and\\Applied Science\\
%University of Pennsylvania\\
%Philadelphia, PA 19104\\
%Email: kumar@seas.upenn.edu}}

%\maketitle

\begin{abstract}
%A key challenge in robot-environment interactions arises from the need to design resilient robots able to survive collisions. 
Collision detection and recovery for aerial robots remain a challenge because of the limited space for sensors and local stability of the flight  controller. We introduce a novel collision-resilient quadrotor that features a compliant arm design to enable free flight while allowing for one passive degree of freedom to absorb shocks. We further propose a novel collision detection and characterization method based on Hall sensors, as well as a new recovery control method to generate and track a smooth trajectory after a collision occurs. Experimental results demonstrate that the robot can detect and recover from high-speed collisions with various obstacles such as walls and poles. Moreover, it can survive collisions that are hard to detect with existing methods based on IMU data and contact models, for example, when colliding with unstructured surfaces, or being hit by a moving obstacle while hovering. 
\end{abstract}

%\begin{keywords}
%Aerial systems: applications, robot safety, collision detection and recovery, field robots.
%\end{keywords}

\section{Introduction}
%\subsection{Motivation}
Unmanned aerial vehicles (UAVs) proliferate across applications~\cite{gupte2012survey}. Among UAVs, the quadrotor has become the standard of practice~\cite{Karydis2017EnergeticsIR}. Advances in control of quadrotors~\cite{lee2010geometric,  mellinger2012trajectory}, allow them to %be involved in tasks of interacting with objects or environments, 
perform tasks such as acrobatics~\cite{brescianini2013quadrocopter},  grasping~\cite{thomas2014toward}, perching~\cite{thomas2015planning}, carrying suspended payloads~\cite{tang2018aggressive}, and flying through narrow gaps~\cite{falanga2018foldable, bucki2019design}. 
An integral part of UAV operation is collision avoidance. 
However, despite advances in integration of perception and planning for collision avoidance (e.g.,~\cite{mohta2018fast, aoudeLJRH13, falanga2020dynamic}), quadrotors still remain vulnerable to collisions that may occur because of unreliable sensors or unpredictable disturbances~\cite{abcd}. The challenge becomes pronounced as quadrotors are tasked to operate in increasingly complex, cluttered and partially-known environments. %, they remain vulnerable to collision due to unreliable sensors or unpredictable disturbances~\cite{abcd}. 

%Although most of the commercial UAVs utilize obstacle avoidance to achieve collision-free trajectories, 

The thesis adopted is this work is that instead of completely avoiding collisions %, which primarily relies on tight integration of perception and planning~\cite{mohta2018fast, aoudeLJRH13, falanga2020dynamic}, 
it may be beneficial to cope with them~\cite{de2005sensorless, karydis2014planning, erez2012trajectory, posa2014direct}. %\todo{more refs} % I removed the CDC because it exceeds the 8 pages limit. 
In fact, sustaining flight after collision---a property that most existing works cannot guarantee~\cite{tomic2017external}---may be crucial to deploy quadrotors in complex and cluttered environments.  %based on the fact that most conventional UAVs are unable to sustain flight after colliding with obstacles. 
The capability to withstand and recover from collisions has been shown to benefit aerial robot navigation in partially-known environments~\cite{mulgaonkar2017robust,mulgaonkar2020tiercel}.

\begin{figure}[t!]
	\vspace{6pt}
	\centering %
	\includegraphics[trim={0 0cm 0cm 0cm},clip,width=0.42\textwidth]{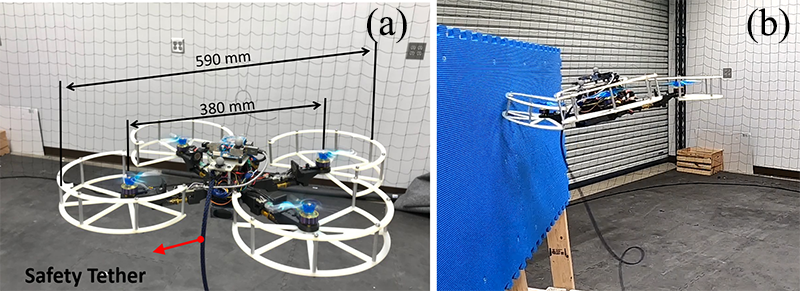}
	%\vspace{-3pt}
	\caption{Snapshot of (a) our new robot colliding (b) with a vertical wall. The supplementary video demonstrates the various types of collision considered herein in a clear way. %Instances of our novel aerial robot detecting and recovering from passive collisions (panels a \& b) and from collision with a vertical wall (panels c \& d). The supplemental video file demonstrates the robot's behavior clearly.} %ARQ can sustain flight after tracking a smooth trajectory for recovery.
	}
	\label{fig:cover}
	\vspace{-18pt}
\end{figure}

We introduce a novel actively resilient quadrotor (ARQ), which incorporates passive springs within its frame to absorb shocks and survive collisions. The quadrotor is equipped with Hall sensors to accurately and rapidly detect the location (in the robot's frame) and intensity of a collision. To take advantage of the injected compliance and resulting resilience to collisions, we further develop a recovery controller that enables the quadrotor to sustain flight after active collision with walls, poles and unstructured obstacles, or after being passively collided by moving objects while hovering (Fig.~\ref{fig:cover}).

%\subsection{Contributions of this Work}
%The goal of this work is to enable aerial robots to survive the impact and sustain controlled flight following a collision. 
%The hypothesis is that integration of passive compliant arms and a recovery control strategy can help quickly identify and mitigate the effect of collisions. 
%Our approach includes 1) the design of a novel compliant and collision-resilient aerial robot prototype called the \emph{actively resilient quadrotor (ARQ)}; 2) the introduction of an algorithm to detect and characterize collisions based on Hall sensors integrated with the robot's frame; and 3) the development of a single-stage collision recovery controller. Evaluation focuses on 1) characterizing the robot's resilience under a range of distinct collisions with walls, poles and unstructured surfaces, passive collisions (i.e. being hit), and free fall tests; 2) quantifying the influence of Hall sensors to the Flight Control Unit; and 3) studying the robot's performance in wall collisions with compliant arms and with recovery control, with compliant arms and no recovery control, and with rigid arms and no recovery control.
%In this work, we introduce a novel active resilient quadrotor (ARQ) and describe both its hardware and software design. 

The ARQ design successfully combines the advantages of two main strategies of existing collision-resilient robots (see Section~\ref{sec:related}) by incorporating passive springs in the frame together with regular rigid protective cages to reduce harmful impacts; these protect the robot from impact speeds in excess of 5.9 m/s.  %Hall sensors along the springs offer fast and accurate detection of a collision. %Thanks to the added compliance and resilience, 
%ARQ is shown able to both survive a wide range of collisions %(including passive collisions and collisions with walls, poles and unstructured obstacles)
%and to sustain flight after recovery.
%
%To the authors' best of knowledge, ARQ is the first compliant quadrotor that can sustain flight after collisions. 
Moreover, ARQ can sustain flight after collision with walls with the fastest speeds to date, and it can recover from active and passive collisions with different obstacles without any prior knowledge on collision models.

The contributions of this work are as follows:
\begin{itemize}
	\item We design, fabricate, and test a novel actively resilient quadrotor that incorporates compliance and sensors to withstand and detect collisions.
	
	\item We propose a collision detection and characterization method based on Hall sensors. 
	
	\item We propose a recovery control method that generates and tracks a smooth trajectory after colliding.  
	
	\item We investigate our robot's collision-resilient capabilities on collisions with walls, poles, unstructured surfaces, and passive collisions (i.e. being hit and free fall).

\end{itemize}

%\todo{remove the outline, optional}The remainder of the paper is organized as follows. Section \ref{design} elaborates on the hardware design, modeling and control, and software architecture of ARQ. Section \ref{collision} describes our method to detect and characterize collisions based on Hall sensors. Section \ref{recover} proposes a novel recovery control strategy for ARQ. Section \ref{experiments} describes the experimental setup and presents the results. \kkb{Section \ref{discussion} discusses our findings and limitations of the current design while Section \ref{conclusion} concludes.}

%%%%%%%%%%%%%%%%%%%%%%%%%%%%%%%%
\section{Related Work}\label{sec:related}
Collision-resilient UAVs are designed to survive crashes with obstacles with no irreversible structural damage~\cite{mintchev2017insect}. So far, research in collision-resilient UAVs has taken two different, largely non-overlapping routes: (1) incorporating protective frames to reduce chances of catastrophic impact, and (2) utilizing sensors to detect and recover from collisions.

Approaches under the first case rely on mechanical design to offer passive protection to the UAV. Cages or protective structures are added to prevent damaging sensitive elements of the robot~\cite{mulgaonkar2017robust,klaptocz2013euler, naldi2014robust}.  Other attempts involve the use of rotation to isolate effects of a collision on the robot~\cite{briod2014collision}, or exploit the added angular momentum to improve disturbance rejection~\cite{abcd}. Another way is to use bio-inspired strategies and soft materials to survive crashes~\cite{sareh2018rotorigami, mintchev2017insect, shu2019quadrotor}. However, careful mechanical design alone cannot help detect and characterize collisions, which is essential to sustain flight.% following a collision.

Works on sensor-based collision detection and characterization have mainly focused on using data from the inertial measurement unit (IMU). External wrench estimation~\cite{tomic2017external}, multiplicative extended Kalman filter (MEKF)~\cite{battiston2019attitude} and fuzzy logic process (FLP)~\cite{dicker2017quadrotor} have been found feasible to detect and characterize collisions with the onboard IMU, to assist recovery control. However, those methods rely on prior knowledge of collision dynamics or require data for training collision models, thereby limiting applicability for collision-resilient control.  Further, IMUs are usually unable to distinguish collisions during aggressive flight and to detect static contacts, resulting in low accuracy in collision detection~\cite{briod2013contact}. Position error between IMU data and ground truth from motion capture~\cite{mulgaonkar2017robust}, and accelerometer data alone~\cite{mulgaonkar2020tiercel} can identify collisions; however, the methods cannot localize collisions, which is key to sustain flight. 
%
%\kkm{Besides IMUs, %there exist other ways to detect failures with sensors. One way 
%another way is to incorporate camera and distance sensors to recover from failures of the vision-based state estimator~\cite{faessler2015automatic}, although it is not possible to handle actual collisions.}\todo{do we need this?we can remove} 

The use of Hall sensors (as in here) has been used in the past to provide more accurate collision detection~\cite{briod2013contact}, albeit for a very different airframe with two rotors aligned vertically. However, and critically different from our work,~\cite{briod2013contact} does not focus on collision recovery control. %thus cannot have the agility of the quadrotor with advanced control strategies~\cite{mellinger2012trajectory, faessler2017differential}. 
In addition, when compared to other works on design of resilient UAVs with collision detection, our work uniquely integrates physical compliance to reduce the detrimental impacts of collision.
%%%%%%%%%%%%%%%%%%%%%%%%%%%%%%%%

\section{Hardware and Software Design of ARQ}\label{design}

%To achieve active collision-resilience while being compliant to reduce detrimental impacts, a quadrotor prototype, named ARQ, has been designed and built. 

\begin{figure}[h!]
 	\vspace{-9pt}
 	\centering %
 	\includegraphics[trim={0cm 0cm 0cm 0.45cm},clip,width=0.40\textwidth]{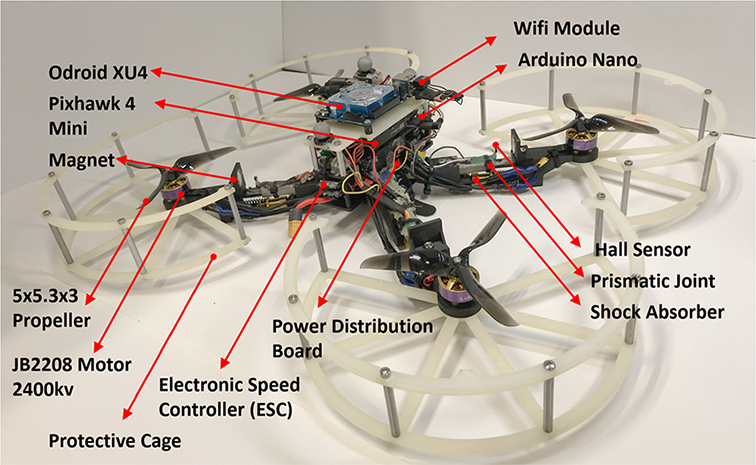}
 	%\vspace{-3pt}
 	\caption{ARQ main components. Each arm is equipped with a shock absorber and Hall sensor to withstand and detect collisions.}
 	\label{fig:drone}
 	\vspace{-16pt}
 \end{figure}

%The focus of this section is on the hardware and software design of ARQ. 
The quadrotor is built based on off-the-shelf components and custom 3D-printed parts. To embrace collision and enable recovery, each arm of ARQ integrates a shock absorber and Hall sensor. The platform features custom nylon protective cages, a Pixhawk 4 Mini flight controller with the open-source PX4 autopilot firmware, an Odroid XU4 with the ROS environment as the companion computer, and an Arduino Nano for analog to digital conversion (Fig.~\ref{fig:drone}).

\subsection{Electro-mechanical Hardware Prototyping}
Key components of ARQ include its compliant arm design and the integration of Hall sensors to estimate deformations.% due to a collision. %The platform features custom nylon protective cages, 3D-printed arms with 5-inch propellers, a Pixhawk 4 Mini flight controller with the open-source PX4 autopilot firmware, an Odroid XU4 with the ROS environment as the companion computer, and an Arduino Nano for analog to digital conversion.

%\subsubsection{Overview}

\subsubsection{Compliant Arm Design}

Unlike existing resilient flying robots whose physical configurations are drastically changed after collision~\cite{shu2019quadrotor, mintchev2017insect}, the configuration of ARQ remains the same before and after a collision. This property makes ARQ an appropriate platform to conduct complex tasks that require model-based control. To this end, our design uniquely retains rigidity when in free flight while allowing for one passive degree of freedom in the direction along each arm to absorb harmful impacts from collision.

\begin{figure}[h!]
	\vspace{-6pt}
	\centering %
	\includegraphics[trim={0cm 0cm 0cm 0cm},clip,width=0.35\textwidth]{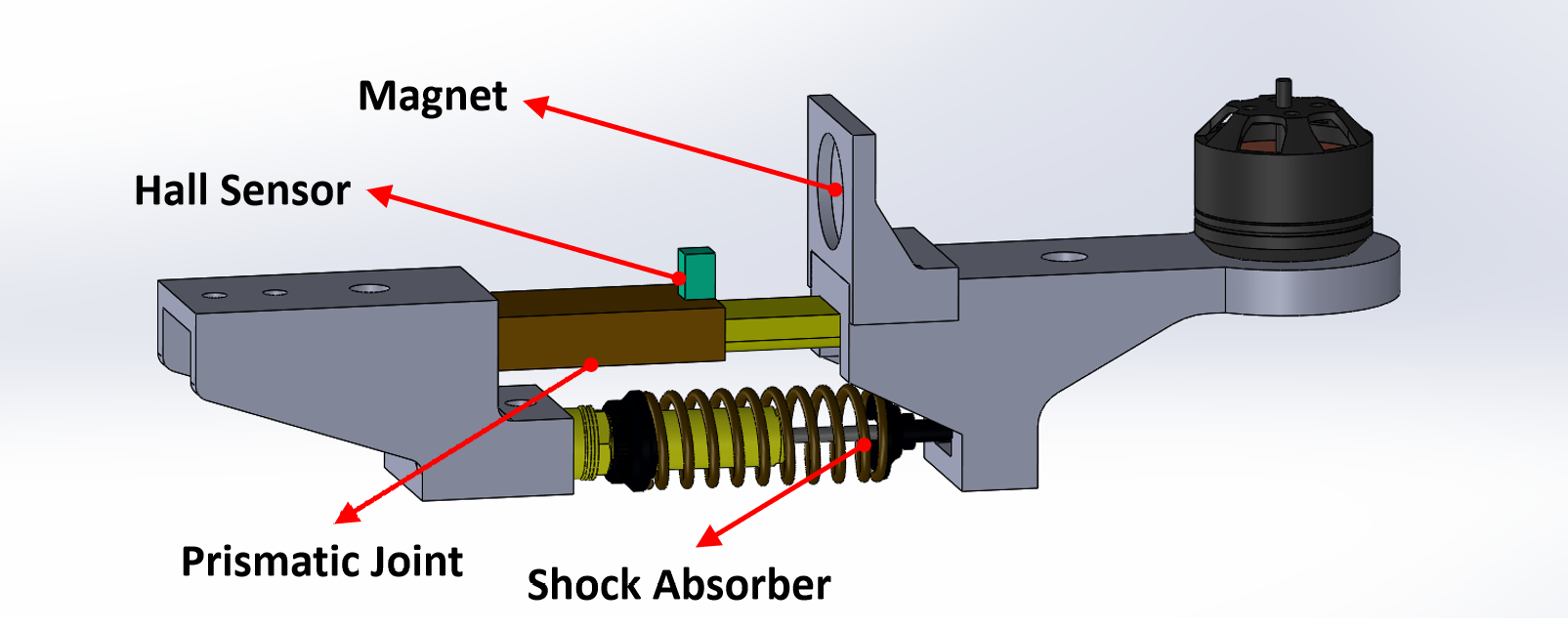}
	\vspace{-3pt}
	\caption{The computer-aided design (CAD) for an ARQ's novel arm.}
	\label{fig:joint_cad}
	\vspace{-6pt}
\end{figure}

The computer-aided design (CAD) assembly and main components of ARQ's novel arm are shown in Fig.~\ref{fig:joint_cad}. The prismatic joint is built based on a solid brass 2-inch surface bolt. The bolt is covered by heat-shrinkable tubes to reduce the free space within the joint. The aluminum shock absorber is taken from 1/18 radio-control cars. An A1302 ratiometric linear Hall sensor is fixed on the prismatic joint to measure the magnetic intensity. The adapters connecting the prismatic joint and shock absorber are 3D-printed (Markforged Mark II, onyx material with carbon fiber add-in).        
The stiffness of the spring within the shock absorber is adjusted so that the length of the arm remains unchanged during free flight. %A close-up view of an arm of ARQ before and during collision is shown in Fig.~\ref{fig:detail}. 
%The novel arm successfully addresses the challenge that the frame remains rigid during regular flight while it can absorb harmful impacts from collision.

% \begin{figure}[h!]
% %	\vspace{-3pt}
% \centering
% 	\begin{subfigure}{.24\textwidth}
% 		\centering
% 		\includegraphics[height=0.75in]{detail1}
% 	\end{subfigure}%
% 	\begin{subfigure}{.24\textwidth}
% 		\centering
% 		\includegraphics[height=0.75in]{detail2}
% 	\end{subfigure}
	
% 	\caption{A close-up view of an ARQ's arm before (left) and during (right) a collision in the direction along the arm. The shock absorber can reduce detrimental impacts and the Hall sensor can detect changes in distance from the magnet.}
% 	\label{fig:detail}
% 	\vspace{-6pt}
% \end{figure}

\subsubsection{Hall Sensor}
The ability to detect and characterize collisions is critical to stabilize quadrotors and sustain flight. Existing collision-detection methods with IMUs require detailed modeling of the environment and obstacles~\cite{dicker2017quadrotor}, limiting application in unknown environments. In this work, we use Hall sensors for collision detection as in~\cite{briod2013contact}.   

Hall sensors are commonly used to measure the magnitude of a magnetic field. The sensor's output voltage is directly proportional to the magnetic field strength through it. In our prototype, A1302 ratiometric linear Hall sensors are fixed on the prismatic joints, coming in contact with the magnet when the length of a shock absorber reduces to its minimum.  %(Fig.~\ref{fig:detail}). 
Despite the relatively short distance, the rotating motors were observed to have almost no effect on the Hall sensors' readings.  Experimental testing showed that the output of the sensor was changed by less than 0.5\% for all motor speeds. Hall sensors and embedded magnets do not affect the IMU in the flight control unit as well, which makes the detection method promising to be adopted outdoors. Experimental testing showed that the yaw estimation of the IMU was not affected by Hall sensors or the embedded magnets.

Admittedly, using only one sensor for each arm does not allow to measure a collision precisely, unless the direction of collision is exactly aligned with the arm. However, it allows to detect whether the quadrotor is in contact with an obstacle or not, and give an approximation of the collision intensity. By utilizing four Hall sensors, the quadrotor can approximate where the collision occurs in the body frame. We elaborate on the collision detection and characterization in Section \ref{collision}.

\subsection{Modeling and Control}

\subsubsection{Notation} 
We use world, $W$, and body, $B$, frames with orthonormal bases $\{\x_W,\y_W, \z_W  \}$ and $\{\x_B,\y_B, \z_B \}$, respectively. The body frame is fixed to the quadrotor with the origin located at the center of mass. % (Fig.~\ref{fig:model}). 
The position, velocity and acceleration of the quadrotor's center of mass are denoted by $\x \in \mathcal{R}^3$, $\vv \in \mathcal{R}^3$ and $\aaa \in \mathcal{R}^3$, respectively. 
Vector subscripts denote the frame the vector is expressed at, for example, $\x_B$ represents the position in the body frame.  Vectors without subscript are expressed in the world frame. $\R^B_W$ denotes a rotation matrix from the body frame to the world frame. For clarity, in the following we use $\R$ to represent $\R^B_W$; subscript $d$ (e.g. $\ddot{\x}_d$) denotes desired values.

%  \begin{figure}[t!]
% 	\vspace{3pt}
% 	\centering %
% 	\includegraphics[trim={0cm 0cm 0cm 0cm},clip,width=0.35\textwidth]{model}
% 	\vspace{-3pt}
% 	\caption{Quadrotor model coordinate systems.}
% 	\label{fig:model}
% 	\vspace{-16pt}
% \end{figure}

\subsubsection{Model} The quadrotor equations of motion are~\cite{mulgaonkar2017robust}
\setlength{\arraycolsep}{0.0em}
\begin{eqnarray*}
\dot{\x}&{}={}&\vv\\
m\dot{\vv}&{}={}&-mg\e_3 + f\R\e_3 \\
\dot{\R}&{}={}&\R\hat{\OO}  \\
J\dot{\OO} + \OO \times J\OO &{}={}&\M
\end{eqnarray*}
where $m$ is the mass and $g$ is the gravitational acceleration constant. Unit vector $\e_3$ represents the direction of gravity in the world frame. $\OO$ denotes the angular velocity in the body frame, and $\hat{\OO}$ is the skew-symmetric matrix representation of $\OO$. $f$ and $\M = \begin{bmatrix}
M_1&\ M_2&\ M_3
\end{bmatrix}^T$ are the force and torque control inputs. These are determined by %The required thrust of each rotor $f_i$ can be determined by solving the system of equations 
\setlength{\arraycolsep}{0.0em}
\begin{eqnarray}\label{force}
\begin{bmatrix}
f\\M_1\\M_2\\M_3
\end{bmatrix}&{}={}&\begingroup % keep the change local
\setlength\arraycolsep{3pt}
\begin{bmatrix}
1&1&1&1\\
0&l&0&-l\\
-l&0&l&0\\
c_{f}&-c_{f}&c_{ f}&-c_{ f}
\end{bmatrix}
\endgroup\begin{bmatrix}
f_1\\ f_2\\ f_3\\ f_4
\end{bmatrix}\enspace,
\end{eqnarray}
where $l$ is the arm length and $c_{ f}$ is the thrust coefficient. 

When a collision happens, the arm length shortens, thus altering the required thrust as per~\eqref{force}. However, by taking advantage of Hall sensors, ARQ waits until an elastic collision has been terminated and the nominal arm length $l$ is recovered. This way, model~\eqref{force} can be used before and after collision.%and hence the system of equations to find the thrust for each motor.
\footnote{During waiting to recover the original arm
length ($<1$\;sec), the flight controller computes rotor thrusts
according to the original arm length $l$.}% and the motors continue rotating. %\todo{not clear what ``hold" means}

\subsubsection{Controller} 
We deploy a geometric tracking controller for quadrotors~\cite{lee2010geometric}. Control inputs $f$, $\M$ are chosen as  
\vspace{-2pt}
\setlength{\arraycolsep}{0.0em}
\begin{eqnarray}
f=&&(-k_x\e_x-k_v\e_v +mg\e_3 + m\ddot{\x}_d)\cdot \R\e_3\\
\M=&&-k_R\e_R -k_\Omega\e_\Omega + \OO\times J\OO \nonumber\\
&&\qquad -J(\hat{\OO}\R^T\R_d\OO_d - \R^T\R_d\dot{\OO}_d)
\end{eqnarray}

Tracking errors are given by
%
% \begin{eqnarray}
% \e_x&{}={}&\x - \x_d\\
% \e_v&{}={}&\vv - \vv_d\\
% \e_R&{}={}&\frac{1}{2}(\R^T_d\R - \R^T\R_d)^\vee\\
% \e_\Omega&{}={}&\OO - \R^T\R_d\OO_d
% \end{eqnarray}
%\begin{eqnarray}
$\e_x=\x - \x_d$; $\e_v=\vv - \vv_d$; $\e_R=\frac{1}{2}(\R^T_d\R - \R^T\R_d)^\vee$; and
$\e_\Omega=\OO - \R^T\R_d\OO_d$, 
where the vee map $^\vee$ is the inverse of a skew-symmetric mapping. Readers are referred to~\cite{lee2010geometric} for detailed derivations and stability analysis of the controller.

%%%%%%%%%%%%%%%%%%%%%%%%%%%%%%%%%%%%
\subsection{Software Architecture}
\begin{figure}[h!]
	\vspace{6pt}
	\centering %
	\includegraphics[trim={0cm 0cm 0cm 0cm},clip,width=0.38\textwidth]{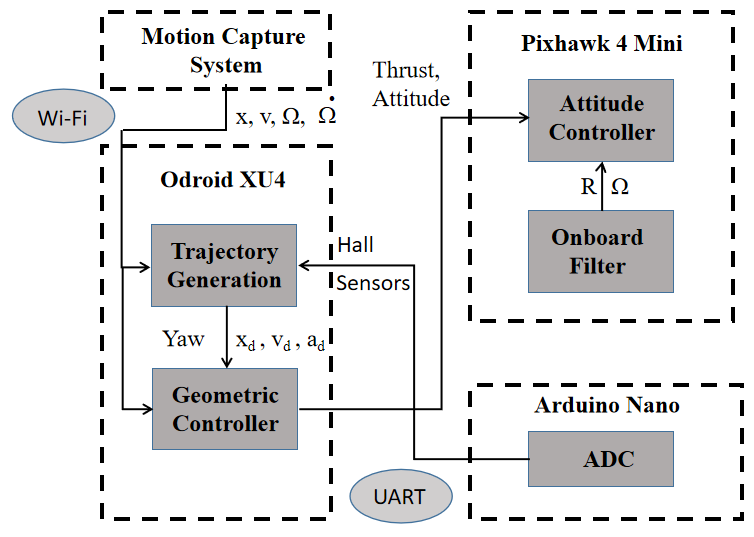}
	\vspace{-3pt}
	\caption{Overview for the software architecture of ARQ.}
	\label{fig:overview}
	\vspace{-9pt}
\end{figure}

The software is developed in C++ using the ROS framework (Fig.~\ref{fig:overview}). The flight controller Pixhawk 4 Mini, companion computer Odroid XU4 and the Arduino Nano are all mounted on the quadrotor (Fig.~\ref{fig:drone}). The Odroid runs Ubuntu 16.04 LTS and communicates with a 12-camera VICON motion capture system server over Wi-Fi for odometry feedback.   

Three nodes are running within the Odroid. The first one is the MAVROS node to communicate with the PX4 firmware in Pixhawk via MavLink. The second node is the trajectory generation node, which will be elaborated in Section \ref{recover}. The trajectory generation node outputs the desired position, velocity, acceleration, yaw, and yaw rate to the geometric controller node. Finally, the controller node calculates and sends the desired thrust and attitude to the flight controller.    

The Pixhawk 4 Mini runs in offboard mode. The inbuilt MAVROS function \texttt{setpoint\_raw/attitude} is called at 50Hz to set motors commands according to the given thrust and attitude data. The Pixhawk runs an internal extended Kalman filter (EKF) to fuse IMU data and estimate the quadrotor's pose and orientation.  

One Arduino Nano connects four Hall sensors and converts the analog signals to digital ones. The Arduino sends digital readings to the Odroid via UART at 200Hz for collision detection and characterization.

%\section{Dynamics modeling and Controller Design}\label{model}
%\subsection{Notation}
%\subsection{Equations of Motion}
%\subsection{Geometric Controller}

%%%%%%%%%%%%%%%%%%%%%%%%%%%%%%%%%%%%%%%
%%%%%%%%%%%%%%%%%%%%%%%%%%%%%%%%%%%%%%%
\section{Collision Detection and Characterization}\label{collision}
The Hall sensor used herein outputs a voltage proportional to the magnetic field strength through it. We fix a magnet at one side of the shock absorber and place the Hall sensor at the other end (see Fig.~\ref{fig:joint_cad}). When a collision happens, the contact force will shorten the spring in the absorber, thereby reducing the distance between a Hall sensor with a magnet, thus increasing the output voltage recorded in the Arduino. 

%It is true that when the collision direction is not aligned with the arm, the Hall sensors cannot exactly measure the collision. The estimated intensity of collision has an upper bound when the shock absorber reaches the minimum length. However, by utilizing four Hall sensors, one on each arm, the ARQ can estimate the intensity and direction of the collision rapidly and accurately.\todo{is this paragraph a repetition?  What is new content here?}

\begin{figure}[h!]
	\vspace{-6pt}
	\centering %
	\includegraphics[trim={0cm 0cm 0cm 0cm},clip,width=0.22\textwidth]{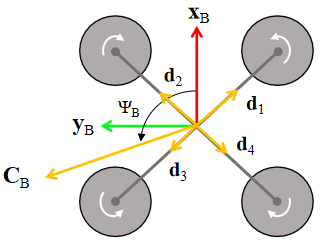}
	\vspace{-3pt}
	\caption{Characterization of a collision vector $\C_B$.}
	\label{fig:collision}
	\vspace{-6pt}
\end{figure}

Figure~\ref{fig:collision} illustrates how ARQ estimates the location and intensity of a collision $\C_B$ in the body frame. %Vectors $\x_B$ and $\y_B$ form an orthonormal basis in the top view. 
There are four vectors $\textbf{d}_i$ that represent the collision detected by the Hall sensors along the arms. Let $d_i \in [0,1], i =1,2,3,4$ denote values from the analog to digital conversions, where $d_i = 0$ means no contact force is detected and $d_i = 1$ indicates that the minimum length of the shock absorber has been reached.

To estimate the intensity and location of the collision $\C_B$, we need to calculate its magnitude $C_B$ and orientation $\Psi_B \in (-\pi, \pi]$ (positive orientation corresponds to counter-clock direction). First, all vectors $\dd_i$ are projected onto the basis $\x_B$ and $\y_B$ with $d_{i,x},d_{i,y}$. For example, the projections of $\dd_1$ can be written as
\setlength{\arraycolsep}{0.0em}
%\begin{eqnarray}
$d_{1,x}=d_1\cos\frac{\theta}{2};\enspace%\\
d_{1,y}=-d_1\sin\frac{\theta}{2}$, 
%\end{eqnarray}
where $\theta = \angle (\dd_1, \dd_2) $ represents the angle between the vectors $\dd_1$ and $\dd_2$. After adding up all projections, we are able to characterize the collision via 
\begin{eqnarray*}
\label{de0}
d_{x}&{}={}&\sum_{i=1}^{4}d_{i,x}\enspace,\\
\label{de1}
d_{y}&{}={}&\sum_{i=1}^{4}d_{i,y}\enspace,\\
\label{de2}
C_B&{}={}&\sqrt{d_x^2+d_y^2}\enspace,\\
\label{de3}
\Psi_B &{}={}&\text{atan2}(d_y,d_x)\enspace.
\end{eqnarray*}
%
% $d_{x}=\sum_{i=1}^{4}d_{i,x}$; 
% $d_{y}=\sum_{i=1}^{4}d_{i,y}$; 
% $C_B=\sqrt{d_x^2+d_y^2}$; and
% $\Psi_B=\text{atan2}(d_y,d_x)$.

An effective collision detection requires both accuracy and an appropriate time of detection $t_d$ to initiate recovery control. Unlike existing methods that often rely on certain thresholds to decide on collision occurrence and detection time, detection herein is based on an algorithm to measure the maximum collision intensity and time $t_d$. %\todo{as I read I feel we should include the algorithm if space allows, and shorten the following paragraph.}  %\kkr{\st{(see Algorithm 1)}}. 

The algorithm initiates with zero for the maximum collision intensity $C_B$ and false for the collision detection flag. While there is no collision detected, the loop reads data from the Hall sensors at current time $j$ and computes $C_{B,j}$ and $\Psi_{B,j}$ as discussed previously. We empirically choose a threshold of 0.1 to reduce effects from sensor noise. When a collision intensity greater than the threshold is detected, the loop compares the current collision magnitude with the maximum one and saves the value if the current collision magnitude is greater. After a maximum collision magnitude is found, the algorithm starts counting for the loop and sets the collision flag true after $N$ loops afterward, where $N$ is selected so that $t_d$ is close to the time when the lengths of the arms recover to satisfy the nominal length dynamic model.

\section{Recovery Control Design}\label{recover}
Existing works in recovery control use different stages to track, resulting in discontinuities in the overall trajectory~\cite{dicker2017quadrotor, dicker2018recovery}. %faessler2015automatic
On the contrary, the proposed recovery controller does not have multiple stages, leading to a smooth desired trajectory to track. This single-stage control design can improve the recovery robustness and applies to collisions with a variety of objects such as walls and poles.

We follow a minimum-snap trajectory generation~\cite{mellinger2011minimum, richter2016polynomial}. Minimum-snap polynomial splines can be effective for quadrotor trajectory planning since the motor commands and attitude accelerations of the vehicle are proportional to the snap of the path. Minimizing the snap of a trajectory can maintain the quality of onboard sensor measurements while avoiding abrupt or excessive control inputs~\cite{richter2016polynomial}.   

After a collision occurs, we seek to make the quadrotor return to a position $\x_d$ that is in the opposite direction of the collision at a distance proportional to the collision intensity. We constrain the recovery position $\x_d$ to be at the same height as the collision position. Thus, we can find the desired position in the body frame $\x_{d,B}$ as
\begin{eqnarray*}
\x_{d,B}&{}={}&\text{Rot}(z, \Psi_B)\begin{footnotesize}\begin{bmatrix}
-C_B\quad 0\quad 0
\end{bmatrix}^T\end{footnotesize}
\end{eqnarray*}  
where $\text{Rot}(z,\Psi_B)$ is the rotation matrix along the $z$-axis by angle $\Psi_B$. Then, $\x_d$ in the world frame is  $\x_d=\R^T\x_{d,B}$.
% \begin{eqnarray}
% \x_d&{}={}&\R^T\x_{d,B}
% \end{eqnarray}  

For the flat output variables $x,y,z$ and yaw angle, we fix the yaw angle and only consider $\x = \begin{bmatrix}
x\quad y\quad z
\end{bmatrix}^T$. A single trajectory segment between two points is composed of independent polynomials $P(t)$ for $\x$. A cost function penalizes the squares of fourth-order derivatives of $P(t)$ as 
 \begin{eqnarray}\label{eq:cost}
J &{}={}& \int_{0}^{T}c_4P^{(4)}(t)^2dt = \p^TQ\p
\end{eqnarray} 
Vector $\p$ contains the $N=10$ coefficients of a single polynomial. Only the fourth-order derivative of $P(t)$ is involved in~\eqref{eq:cost} to minimize snap. We refer readers to~\cite{richter2016polynomial} for details on the construction of the Hessian matrix $Q$.

The optimization is formulated as
\begin{eqnarray*}
\underset{\p}{\text{minimize }} &{}J={}&\p^TQ\p \nonumber\\
%\text{subject to } &{}\x(0) = {}&\x_0, \dot{\x}(0) = \textbf{0},\ddot{\x}(0) = \textbf{0}  \nonumber\\
%					&{}\x(T) = {}&\x_d, \dot{\x}(T) = \textbf{0},\ddot{\x}(T) = \textbf{0}
\text{subject to }&{}A\p = {}&\textbf{b}
\end{eqnarray*}
Equality $A\p = \textbf{b}$ imposes endpoint constraints $\x(0) = \x_0, \dot{\x}(0) = \vv_0,  \x(T) = \x_d$, where $\x_0, \vv_0$ are the position and velocity of the quadrotor when collision happens and $T$ is estimated based on the maximum acceleration and velocity~\cite{richter2016polynomial}. The velocity and acceleration at $t=T$ are set to zero. Note we do not command the acceleration at $t=0$ since feedback from motion capture has no acceleration data.

\section{Experiments and Results}\label{experiments}

%This section presents the experimental setup and results of collision tests for ARQ. 
%The collision-resilient performance of ARQ is evaluated through five experiments: passive collision, wall collision, pole collision, colliding with unstructured surfaces, and free fall.

The ARQ measures 380 mm from propeller tip to tip and 590 mm from protection cage tip to tip. We use a 3500 mAh 3-cell LiPo battery to power the quadrotor. The robot weighs 1.419 kg with the battery and has a maximum thrust-to-weight ratio of 4.58. The average energy efficiency is 2.2 g/W. During experiments, the actual thrust-to-weight ratio is limited to 2 by reducing the maximum pulse width modulation (PWM) values. %A 12-camera VICON motion capture system is used to provide odometry feedback. 
A safety tether connects the robot to prevent damages in case of losing control. %(Fig.~\ref{fig:flight}). 
The safety tether is lightweight, and we did not observe any impact on the flight. We manually command the robot to take off in stabilized mode then switch to the offboard mode to let it hover before tests. When tests are done, we switch back to stabilized mode again to manually land.  %\kkr{\st{We set $N =5$ and threshold of 0.1 in Algorithm 1.} } 
The collision-resilient performance of ARQ is evaluated through five experiments: passive collision, wall collision, pole collision, colliding with unstructured surfaces, and free fall.

%  \begin{figure}[h!]
% 	\vspace{-6pt}
% 	\centering %
% 	\includegraphics[trim={0.7cm 0.1cm 0.8cm 1.1cm},clip,width=0.32\textwidth]{flight}
% 	%\vspace{-3pt}
% 	\caption{ARQ is hovering with a safety tether.}
% 	\label{fig:flight}
% 	\vspace{-9pt}
% \end{figure}

%%%%%%%%%%%%%%%%%%%%%%%%%%%%%%%%%%%%%
\subsection{Passive Collision}
%Existing works on collision detection and recovery assume that the quadrotor operates within a \kkm{static environment}.\todo{ref} 
%In other words, the robot is the only moving object within the environment. 
%However, as UAVs are tasked to explore more dynamic environments, the robot can be hit by an obstacle when it is hovering. We call this type of collision as a \emph{passive collision}. Surviving passive collisions is challenging because such type of collision is almost undetectable by using IMU data alone. There is no change in IMU readings before the collision happens, therefore all IMU-based methods will fail in this case~\cite{tomic2017external, battiston2019attitude, dicker2017quadrotor}. Similarly, vision-based resilience is very likely to fail as well unless the colliding object is right inside the camera field of view and can be detected fast enough. 

\begin{figure}[h!]
	\vspace{-20pt}
	\centering %
	\includegraphics[trim={0 0 0 0.55cm},clip,width=0.40\textwidth]{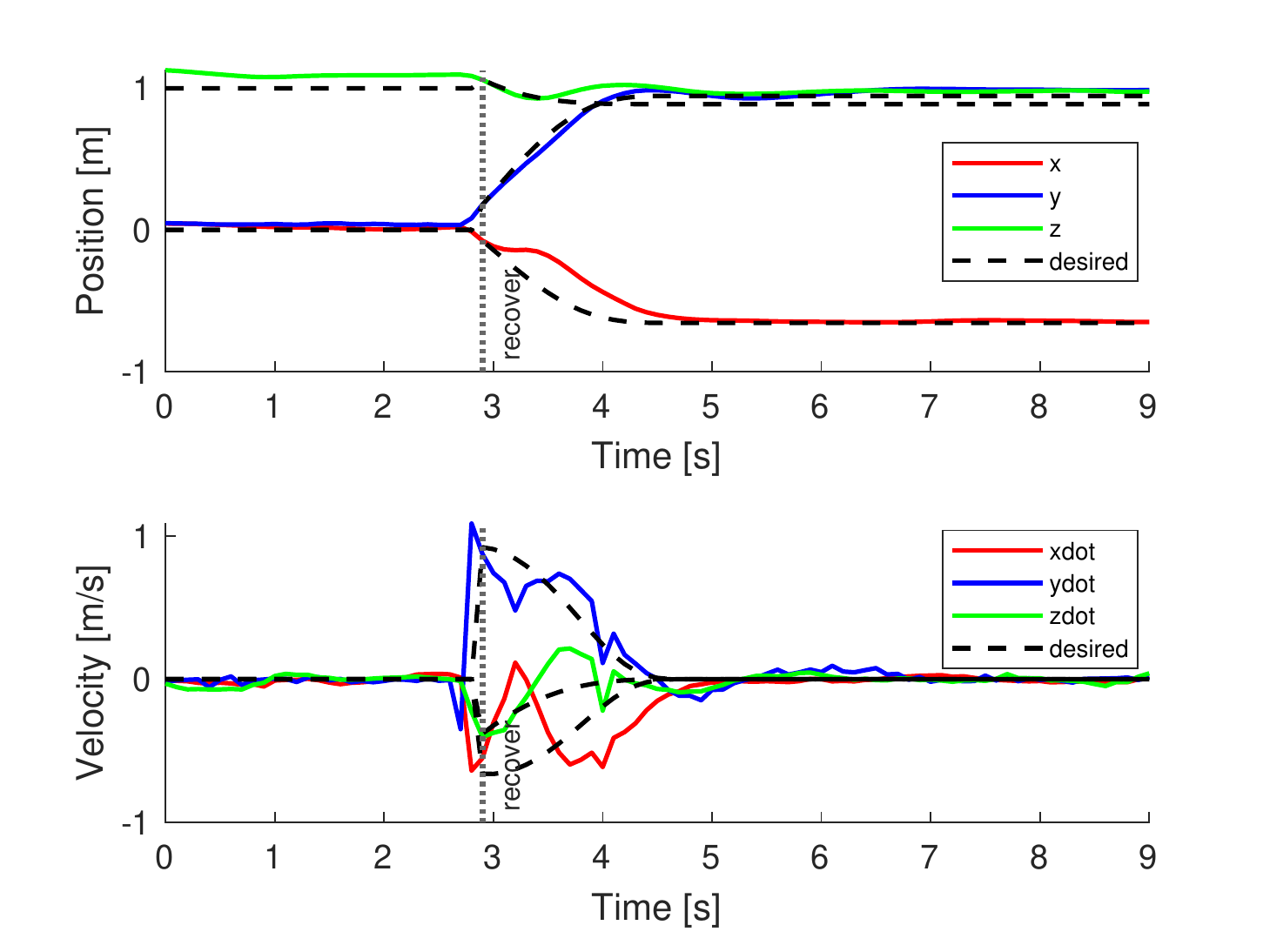}
	\vspace{-3pt}
	\caption{State tracking in passive collision recovery where the $x$-axis points to the North and the $y$-axis points to the West. The origin is located at the ARQ's takeoff position.}
	\label{fig:pp2}
	\vspace{-6pt}
\end{figure}

As UAVs are tasked to explore more dynamic environments, the robot can be hit by an obstacle while hovering. We call this type of collision as a \emph{passive collision}. 
ARQ can handle passive collisions thanks to its equipped Hall sensors that allow for rapid collision detection while hovering. In our tests, the ARQ is being hit along the direction of an arm when it is hovering. %\kkm{Snapshots of the recovery process are shown in Fig.~\ref{fig:pp1}.}
We track actual and desired states during the passive collision recovery in Fig.~\ref{fig:pp2}. %where the positions and velocities in the axis $x$ and $y$ are plotted in red and blue, respectively. 
Desired values are plotted in black dashed curves. A vertical dotted line represents the time when the recovery controller is triggered.

\begin{figure}[!h]
	\vspace{-6pt}
	\centering %
	\includegraphics[width=0.40\textwidth]{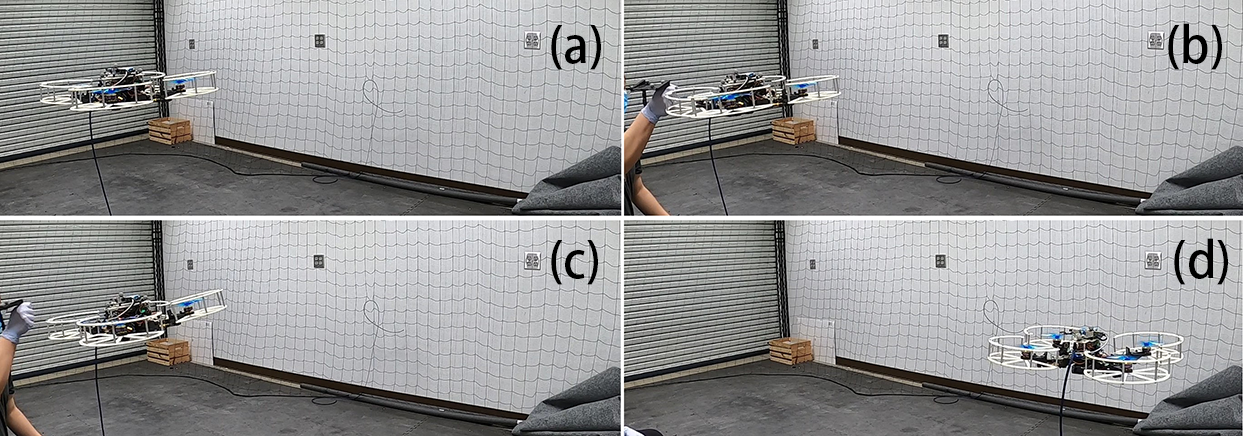}
	\vspace{-0pt}
	\caption{Snapshots of ARQ detecting and recovering from a passive collision. (a) ARQ hovers. (b) Collision starts and the arm absorbs the shock. (c) Recovery control starts and there is a passive body morphing in the air. (d) ARQ is stabilized and hovering again.}
	\label{fig:pp1}
	\vspace{-12pt}
\end{figure}

Results demonstrate that ARQ can sustain flight after being passively hit at a collision speed of 1.3 m/s (i.e. the norm of the vehicle's linear velocity after the collision). 
%
%Figure~\ref{fig:pp2} indicates sudden changes in velocities due to the passive collision. 
After the collision is detected and characterized, ARQ computes a smooth stabilizing trajectory, the initial velocities of which match the ones when recovery control begins. The geometric controller commands ARQ to track the desired trajectory. However, irregular changes in velocities are observed due to the body morphing (Fig.~\ref{fig:pp1}.c). Body morphing delays trajectory tracking in the $x$ direction; however, ARQ can still converge to desired states quickly and sustain stable flight. Low tracking errors in Fig.~\ref{fig:pp2} support the claim that ARQ's arms remain rigid to the extent possible during free flight.

%Snapshots of the recovery process are shown in Fig.~\ref{fig:pp1}.

%%%%%%%%%%%%%%%%%%%%%%%%%%%%%
\subsection{Wall Collision}
%Walls and poles are the most predominant obstacles to deploy UAVs in urban settings~\cite{dicker2018recovery} and all related work is focusing on these obstacles. In this paper, we conduct similar tests to validate ARQ's capability to recover from a collision with a vertical wall and a pole. \todo{maybe remove or push elsewhere?}

In the wall collision test, we command ARQ to follow a trajectory that intersects with the wall. The position of the wall is entirely unknown to the robot before the collision. We consider single-arm and two-arm collision types, and repeat multiple trials to evaluate consistency of collision recovery. %we place a vertical wooden wall covered by foam sheet in front of the robot (Fig.~\ref{fig:wall}). We follow the same steps to take off and hover with the geometric controller. Then we command ARQ to follow a trajectory that intersects with the wall. The position of the wall is entirely unknown to the robot before the collision.

\begin{figure}[h!]
\vspace{6pt}
\centering
	\begin{subfigure}{.45\textwidth}
		\centering
		\includegraphics[trim={0 0 0 0.55cm},clip,width=0.90\textwidth]{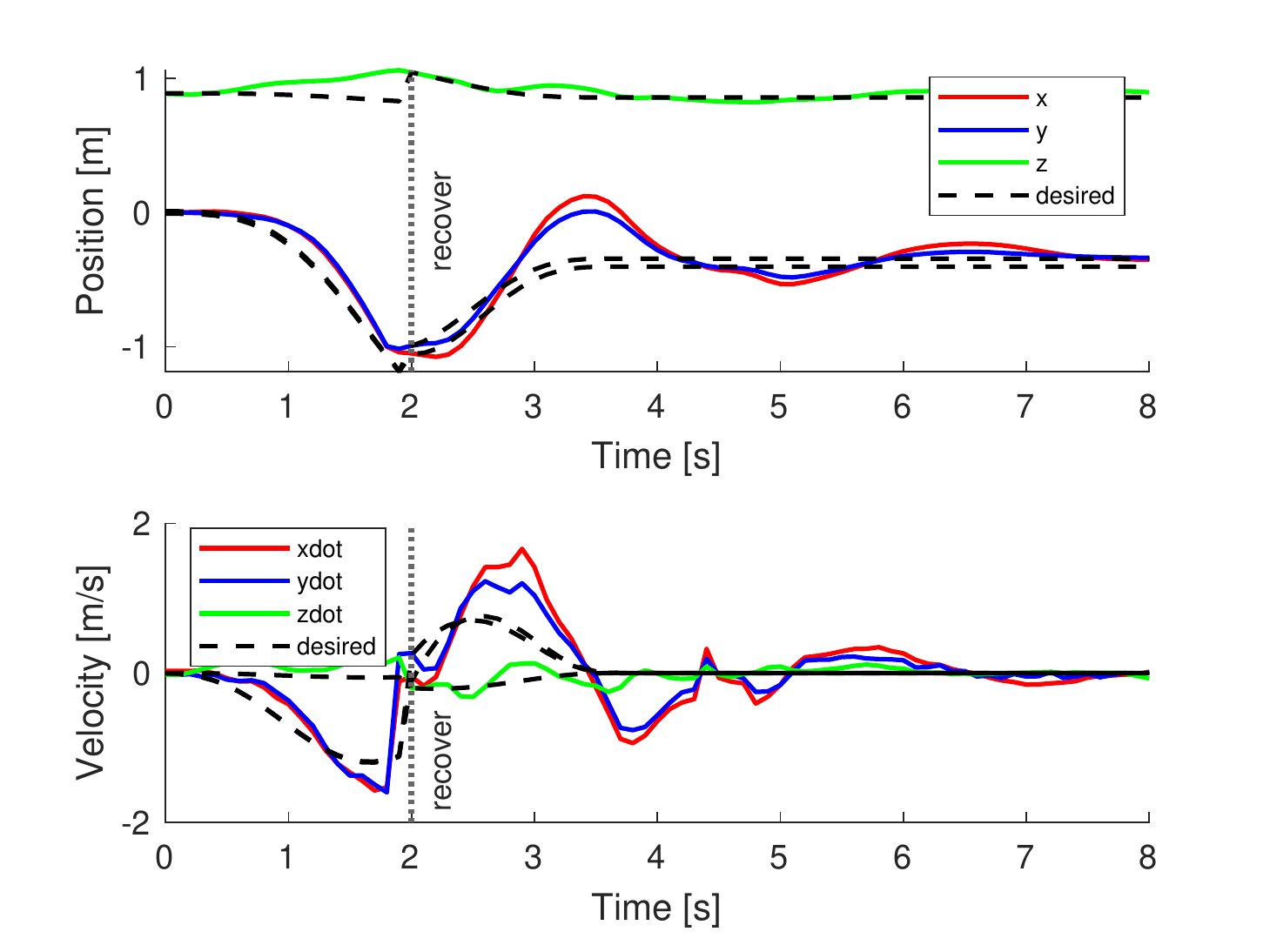}
		\vspace{-6pt}
		 \caption{Single-arm wall collision.}
	\end{subfigure}%
	\vspace{-0pt}
	\begin{subfigure}{.45\textwidth}
		\centering
		\includegraphics[trim={0 0 0 0.55cm},clip,width=0.90\textwidth]{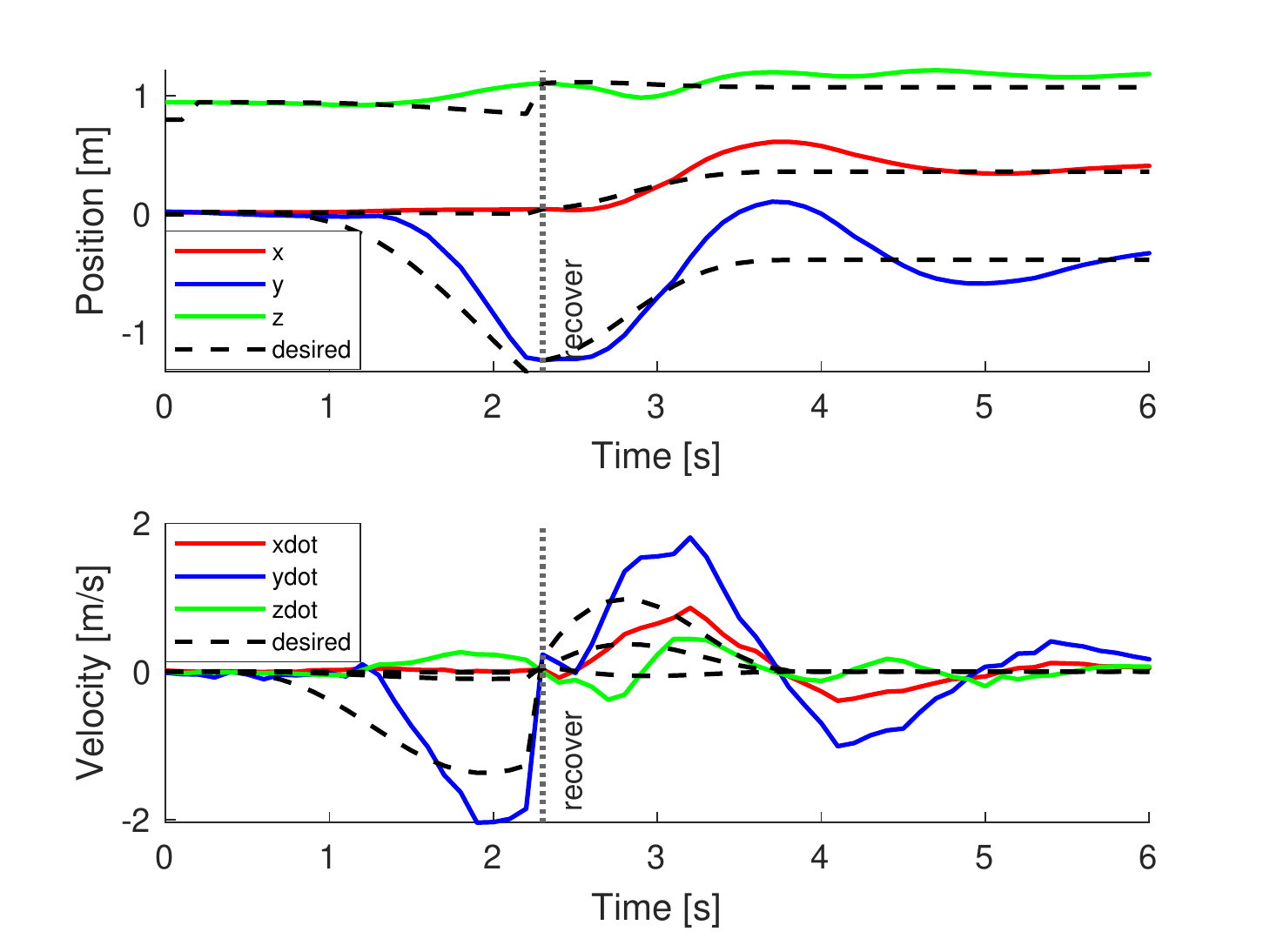}
		\vspace{-6pt}
	 \caption{Two-arm wall collision.}
	\end{subfigure}
	\vspace{-3pt}
	\caption{State tracking during wall collision recovery in two sample cases for (a) single-arm and (b) two-arm collision types.}
	\label{fig:wall_track}
	\vspace{-3pt}
\end{figure}

\begin{figure}[ht!]
	\vspace{-6pt}
	\centering %
	\includegraphics[width=0.40\textwidth]{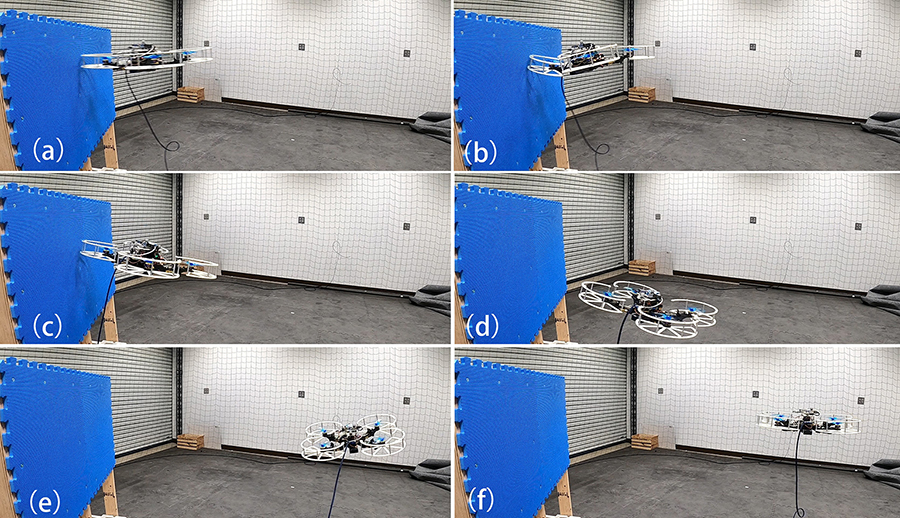}
	\vspace{-2pt}
	\caption{ARQ is recovering from wall collision with a single arm in contact at a collision speed of 2.58 m/s and a collision intensity of 0.97.  %0.9687.
	(a) Collision starts at t = 0 ms. (b) The arm absorbs the shock and the robot comes to a full stop at t = 32 ms. (c) to (e) ARQ is tracking a smooth trajectory for recovery.  (f) ARQ is stabilized and hovering again at t = 3,816 ms (see supplementary video).}
	\label{fig:wall}
	\vspace{-18pt}
\end{figure}

Figure~\ref{fig:wall_track} shows sample trajectories of single- and two-arm wall collision. Figure~\ref{fig:wall} highlights snapshots of a sample single-arm wall collision trial. %(see supplementary video for more sample trials).  
Testing reveals that ARQ can detect the wall accurately and quickly, then recover from collision and sustain flight afterwards with a speed of up to 2.58 m/s, which is greater than previously-reported maximum feasible velocities of 1.0 m/s in~\cite{tomic2017external} and 2.0 m/s in~\cite{dicker2017quadrotor}.  %\kkr{\st{ARQ has the potential to achieve successful recovery at higher collision speeds since it includes physical compliance to absorb shocks (Fig.~11.b) and tracks a smooth trajectory for recovery (Fig.11.c-f). } } 
%We investigate collision resilience to a vertical wall with two arms in contact as well \kkb{(see supplementary video). %\kkr{ \st{Figure  depicts ARQ before and during a collision with a single arm and two arms in contact.}} 
Results validate that ARQ can also estimate the intensity and orientation of two-arm collision accurately, to sustain stable flight with a collision speed of 1.92 m/s. %to match the speed in the fig 9, this is not the highest speeds but more convincing.

We evaluate consistency of collision recovery by executing 20 consecutive wall collision tests in both single-arm and two-arm types (10 in each case; see supplementary video). ARQ had 9/10 successful recoveries (survive collisions and sustain flight) for the two-arm type and 8/10 successful recoveries for the single-arm type at the highest collision speeds, respectively. The robot failed to detect the collision for the only failure in two-arm collision tests, while the ARQ did detect the collision but was unable to stabilize itself due to large inclination angles in the unsuccessful single-arm recovery.

To demonstrate the individual contribution of compliance and the recovery control components, two additional wall tests are conducted with a collision speed of 2.3 m/s. In the first test, the arms remain compliant but the recovery controller is disabled. In the second test, we fix the arms in place using custom jigs (these add 25 g of weight to the robot) while the recovery controller is disabled. Experimental trials (see supplementary video) show that compliance alone can reduce collision impact (as expected), but cannot help sustain fight; the robot falls to the ground after getting
stuck to the wall for several seconds because of lack of recovery control. In the second test, we observe a worse impact which makes the robot fall to the ground immediately. The
two tests support our hypothesis that compliance can reduce detrimental impacts from collisions while recovery control is essential to sustain fight afterwards. %\todo{simplify the paragraph and add success rate experiment results}
% \begin{figure}[t!]
% 	\vspace{6pt}
% 	\centering %
% 	\includegraphics[width=0.48\textwidth]{wall2}
% 	%\vspace{-3pt}
% 	\caption{Snapshots of ARQ detecting and recovering from a vertical wall collision with a single arm (a \& b) and two arms in contact (c \& d).}
% 	\label{fig:wall2}
% 	\vspace{-8pt}
% \end{figure}

\subsection{Pole Collision}

\begin{figure}[h!]
	\vspace{-6pt}
	\centering %
	\includegraphics[trim={0 0 0 0},clip,width=0.40\textwidth]{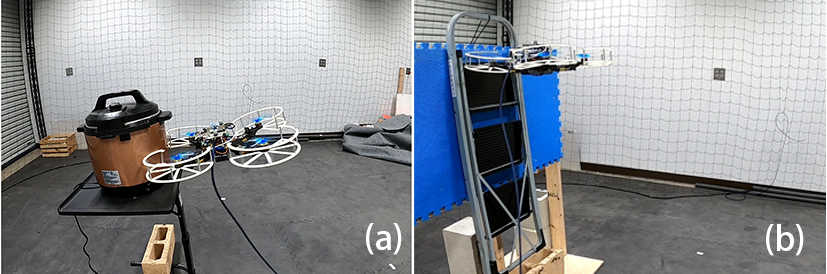}
	%\vspace{-3pt}
	\caption{ARQ can detect and recover from collisions with (a) a pole and (b) unstructured surface (see supplementary video). %collision with the speed of 2.04 m/s and a collision intensity of 0.20.}
	}
	\label{fig:pole}
	\vspace{-6pt}
\end{figure}

The robot can also recover from collisions with cylindrical/pole-like objects. %for the pole collision test. 
In our tests, we use a cylinder of 300 mm diameter (Fig.~\ref{fig:pole}.a) %Figure~\ref{fig:pole}.a shows snapshots when colliding at a speed of 2.04 m/s.
and command ARQ to hit the object with a collision speed of 2.04 m/s. 

Testing demonstrates that ARQ can accurately detect pole collisions without any additional changes in hardware and software. The recovery process is very similar to the one with vertical walls, except the collision intensity (0.20)  %(0.1919) 
is much smaller, due to the shape of the obstacle. However, the recovery controller still manages to stabilize the robot with a hovering position close to the obstacle. %(see Fig.~\ref{fig:pole}.b). 
The recovery does not have multiple discontinuous phases as in~\cite{dicker2018recovery}, and the robot tracks a smooth trajectory throughout the recovery.  

% \begin{figure}[h!]
% 	\vspace{6pt}
% 	\centering %
% 	\includegraphics[width=0.4\textwidth]{un}
% 	%\vspace{-3pt}
% 	\caption{ARQ is detecting and recovering from collision with an unstructured surface with a speed of 1.95 m/s. }
% 	\label{fig:un}
% 	\vspace{-15pt}
% \end{figure} 
%%%%%%%%%%%%%%%%%%%%%%%%%%%%%%%%%
\subsection{Unstructured Surface Collision}

Unlike IMU-based methods, ARQ does not require prior knowledge of collision models to handle collisions. We design a collision test on an unstructured surface (Fig.~\ref{fig:pole}.b) to demonstrate this property. %We place a metal folding step ladder along the wall (see Fig.~\ref{fig:pole}.b). 
It is very challenging to derive a contact model for such irregular surfaces, while the collision position is uncertain that makes it even harder to detect the collision based on contact modeling.

We command ARQ to hit the unstructured surface with a collision speed of $1.95$\,m/s. %The process is illustrated in Fig.~\ref{fig:pole} (b). 
Testing demonstrates that ARQ can quickly and accurately detect collisions with unstructured surfaces thanks to the Hall sensors along the arms. Following a smooth desired trajectory after the collision, ARQ can stabilize itself and return to the hovering state. ARQ's ability to survive a collision with the unstructured surface is key for UAVs to be deployed in unknown environments.

\subsection{Free Fall} 
\begin{figure}[h!]
\vspace{-6pt}
\includegraphics[width=0.325\textwidth]{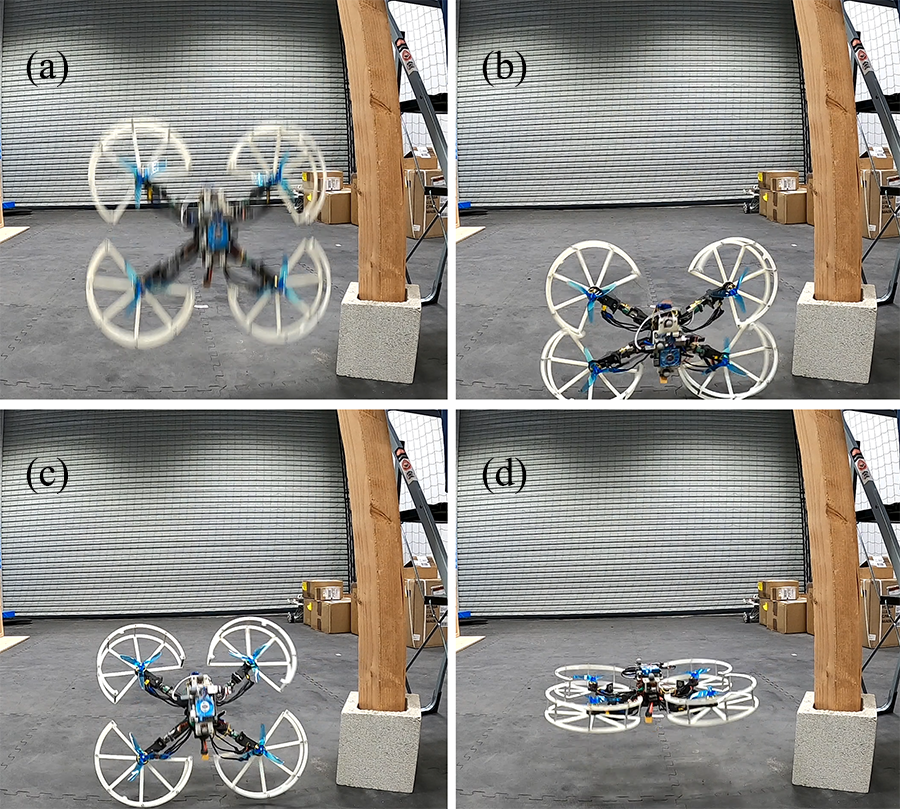}
\centering
\caption{Drop test from $1.8$\,m reaching a velocity of $5.9$\,m/s. }
\label{fig:drop}
\vspace{-12pt}
\end{figure}

To better demonstrate the benefit of adding compliance, we conduct a free fall test with ARQ (Fig. \ref{fig:drop}). ARQ is found to survive free falls from 1.8 m high and reaching a maximum velocity of 5.9 m/s without any damage to the robot.\footnote{The battery is removed in this test as the physical
connection in this first prototype is loose, but future iterations of the robot will resolve this; however, removing the battery does not alter the nature of the free fall test. The robot weighs $1.12$\,kg without the battery.}

%%%%%%%%%%%%%%%%%%%%%%%%%%%%%
%%%%%%%%%%%%%%%%%%%%%%%%%%%%%
%\section{Discussion}\label{discussion}
%\kkb{Despite the robot's collision-resilient performance demonstrated in this work, the ARQ's current design is unable to detect and react to perpendicular obstacles (e.g., collisions from the top or the bottom) or purely rotational impacts. However, the paper centers on collisions resilience against vertical obstacles such as walls and poles, because they are the most predominant obstacles for deploying UAVs in urban settings~\cite{dicker2017quadrotor}. 
%
%On the other hand, we can further avoid collisions perpendicular to the arms by limiting the operating height for the navigation in unknown environments. The proposed configuration is also suitable to incorporate sensors that are required for mapping-based methods. We found our protective cage design in accordance with robots operating in GPS-denied environments with cameras and laser scanners~\cite{mohta2018fast}.} 

%\kkb{We observed a 100\% accuracy in at least 50 diverse experimental trials done at different times, which offers preliminary evidence that the collision detection based on Hall sensors is accurate. Nevertheless, a more comprehensive study with a larger number of repeated experiments would further strengthen this evidence.}\todo{remove the paragraph} 

\section{Conclusions}
\paragraph*{\textbf{Contributions and Key Findings}} 
The paper introduces a novel actively resilient quadrotor (ARQ) that can detect collisions quickly and accurately while staying compliant to mitigate damaging collision impacts.  %\kkr{\st{ Critical to this is the quadrotor's novel compliant arm design. } } 
The arm is shown to successfully meet the challenge of retaining rigidity in the absence of collision (i.e. during free flight) while allowing one passive degree of freedom in the direction along the arm. Four Hall sensors with shock absorbers make ARQ able to detect collision without IMU data and prior modeling information for obstacles. With the extra resilience to collisions, we propose a new collision detection and characterization method, followed by a novel recovery control method to generate and track a smooth trajectory for collision recovery.  

Experimental results demonstrate that ARQ can detect and recover from several types of collision at high speeds.  %\kkr{\st{Specifically, we consider collisions with walls and poles that are the most predominant obstacles to be met when deploying UAVs in urban settings, and with irregular objects that may be found in more general environments (e.g., a warehouse).} } 
We further study passive collisions (i.e. the robot being hit by a moving obstacle), which may occur unexpectedly at dynamic environments. 
In all cases, the robot can %\kkr{\st{ detect, handle and recover flight after collisions} }
recover from collisions and sustain flight afterwards. ARQ had 80\% success rate for single-arm wall collisions at a speed of 2.58\,m/s and 90\% for two-arm wall collisions at 1.92\,m/s. %Moreover, thanks to the additional resilience based on Hall sensors, ARQ can detect and recover from challenging collision tasks such as passive collision and unstructured surface collision, which are hard to detect due to lack of changes in IMU data or prior knowledge on the modeling of obstacles.  

%Future work includes a comprehensive study for feasible maximum velocities for collision with various obstacles. We plan to conduct fatigue tests to measure the success rate for the collision. \kkb{We will explore the scalability and its effect on collision-resilient performance of the robot.} We also seek to deploy ARQ in outdoor environments with navigation tasks without obstacle avoidance. 

\paragraph*{\textbf{Directions for Future Work}} 
Despite the robot's collision-resilient performance demonstrated herein across objects that are predominantly encountered in applications~\cite{dicker2017quadrotor}, the current design is unable to detect and react to perpendicular obstacles (e.g., collisions from the top or the bottom) or purely rotational impacts. %However, the paper centers on collisions resilience against vertical obstacles such as walls and poles, because they are the most predominant obstacles for deploying UAVs in urban settings~\cite{dicker2017quadrotor}. 
%
%We can avoid collisions perpendicular to the arms by limiting the operating height for the navigation in unknown environments. The proposed configuration is also suitable to incorporate sensors that are required for mapping-based methods. We found our protective cage design in accordance with robots operating in GPS-denied environments with cameras and laser scanners~\cite{mohta2018fast}.
%
Future directions of research include 1) improved design of the protective cage to handle perpendicular to the frame and purely rotational collisions and to adapt to collisions with large inclination angles, 2) a comprehensive study to identify feasible maximum velocities for collision with a larger set of obstacles, %We plan to conduct fatigue tests to measure the success rate for the collision. %\kkb{We will explore the scalability and its effect on collision-resilient performance of the robot.} 
and 3) deployment and testing in outdoor natural environments. %with navigation tasks without obstacle avoidance. 

%Walls and poles are the most predominant obstacles to deploy UAVs in urban settings~\cite{dicker2018recovery} and all related work is focusing on these obstacles. In this paper, we conduct similar tests to validate ARQ's capability to recover from a collision with a vertical wall and a pole. \todo{maybe remove or push elsewhere?}

%\section*{Acknowledgment}
%We gratefully acknowledge the support of NSF under grant \#  IIS-1910087 and of the UCR Office of Research and Economic Development under a Collaborative Seed Grant.  Any opinions, findings, and conclusions or recommendations expressed in this material are those of the authors and do not necessarily reflect the views of the funding agencies.

\balance
%\section*{Acknowledgment}

%========================================================
%========================================================
%\bibliographystyle{apacite}
%\bibliographystyle{vancouver}
%\bibliographystyle{harvard}
\bibliographystyle{IEEEtran}
%\bibliographystyle{plainnat}

%\bibliographystyle{spbasic} 
%\balance
\bibliography{ZLIU_CASE20}

\begin{thebibliography}{10}
\providecommand{\url}[1]{#1}
\csname url@rmstyle\endcsname
\providecommand{\newblock}{\relax}
\providecommand{\bibinfo}[2]{#2}
\providecommand\BIBentrySTDinterwordspacing{\spaceskip=0pt\relax}
\providecommand\BIBentryALTinterwordstretchfactor{4}
\providecommand\BIBentryALTinterwordspacing{\spaceskip=\fontdimen2\font plus
\BIBentryALTinterwordstretchfactor\fontdimen3\font minus
  \fontdimen4\font\relax}
\providecommand\BIBforeignlanguage[2]{{%
\expandafter\ifx\csname l@#1\endcsname\relax
\typeout{** WARNING: IEEEtran.bst: No hyphenation pattern has been}%
\typeout{** loaded for the language `#1'. Using the pattern for}%
\typeout{** the default language instead.}%
\else
\language=\csname l@#1\endcsname
\fi
#2}}

\bibitem{gupte2012survey}
S.~Gupte, P.~I.~T. Mohandas, and J.~M. Conrad, ``A survey of quadrotor unmanned
  aerial vehicles,'' in \emph{Proceedings of IEEE Southeastcon}, 2012, pp.
  1--6.

\bibitem{Karydis2017EnergeticsIR}
K.~Karydis and V.~Kumar, ``Energetics in robotic flight at small scales,''
  \emph{Interface Focus}, vol.~7, 2017.

\bibitem{lee2010geometric}
T.~Lee, M.~Leok, and N.~H. McClamroch, ``Geometric tracking control of a
  quadrotor uav on se (3),'' in \emph{IEEE Conference on Decision and Control
  (CDC)}, 2010, pp. 5420--5425.

\bibitem{mellinger2012trajectory}
D.~Mellinger, N.~Michael, and V.~Kumar, ``Trajectory generation and control for
  precise aggressive maneuvers with quadrotors,'' \emph{The International
  Journal of Robotics Research}, vol.~31, no.~5, pp. 664--674, 2012.

\bibitem{brescianini2013quadrocopter}
D.~Brescianini, M.~Hehn, and R.~D'Andrea, ``Quadrocopter pole acrobatics,'' in
  \emph{IEEE/RSJ International Conference on Intelligent Robots and Systems
  (IROS)}, 2013, pp. 3472--3479.

\bibitem{thomas2014toward}
J.~Thomas, G.~Loianno, J.~Polin, K.~Sreenath, and V.~Kumar, ``Toward autonomous
  avian-inspired grasping for micro aerial vehicles,'' \emph{IOP Bioinspiration
  \& Biomimetics}, vol.~9, no.~2, p. 025010, 2014.

\bibitem{thomas2015planning}
J.~Thomas, G.~Loianno, M.~Pope, E.~W. Hawkes, M.~A. Estrada, H.~Jiang, M.~R.
  Cutkosky, and V.~Kumar, ``Planning and control of aggressive maneuvers for
  perching on inclined and vertical surfaces,'' in \emph{ASME International
  Design Engineering Technical Conferences and Computers and Information in
  Engineering Conference}, 2015.

\bibitem{tang2018aggressive}
S.~Tang, V.~W{\"u}est, and V.~Kumar, ``Aggressive flight with suspended
  payloads using vision-based control,'' \emph{IEEE Robotics and Automation
  Letters}, vol.~3, no.~2, pp. 1152--1159, 2018.

\bibitem{falanga2018foldable}
D.~Falanga, K.~Kleber, S.~Mintchev, D.~Floreano, and D.~Scaramuzza, ``The
  foldable drone: A morphing quadrotor that can squeeze and fly,'' \emph{IEEE
  Robotics and Automation Letters}, vol.~4, no.~2, pp. 209--216, 2018.

\bibitem{bucki2019design}
N.~Bucki and M.~W. Mueller, ``Design and control of a passively morphing
  quadcopter,'' in \emph{IEEE International Conference on Robotics and
  Automation (ICRA)}, 2019, pp. 9116--9122.

\bibitem{mohta2018fast}
K.~Mohta, M.~Watterson, Y.~Mulgaonkar, S.~Liu, C.~Qu, A.~Makineni, K.~Saulnier,
  K.~Sun, A.~Zhu, J.~Delmerico, \emph{et~al.}, ``Fast, autonomous flight in
  gps-denied and cluttered environments,'' \emph{Journal of Field Robotics},
  vol.~35, no.~1, pp. 101--120, 2018.

\bibitem{aoudeLJRH13}
G.~Aoude, B.~Luders, J.~Joseph, N.~Roy, and J.~P. How, ``Probabilistically safe
  motion planning to avoid dynamic obstacles with uncertain motion patterns,''
  \emph{Auton. Robots}, vol.~35, no.~1, pp. 51--76, 2013.

\bibitem{falanga2020dynamic}
D.~Falanga, K.~Kleber, and D.~Scaramuzza, ``Dynamic obstacle avoidance for
  quadrotors with event cameras,'' \emph{Science Robotics}, vol.~5, no.~40,
  2020.

\bibitem{abcd}
N.~{Bucki} and M.~W. {Mueller}, ``Improved quadcopter disturbance rejection
  using added angular momentum,'' in \emph{IEEE/RSJ International Conference on
  Intelligent Robots and Systems (IROS)}, Oct 2018, pp. 4164--4170.

\bibitem{de2005sensorless}
A.~De~Luca and R.~Mattone, ``Sensorless robot collision detection and hybrid
  force/motion control,'' in \emph{IEEE international conference on robotics
  and automation (ICRA)}, 2005, pp. 999--1004.

\bibitem{karydis2014planning}
K.~Karydis, D.~Zarrouk, I.~Poulakakis, R.~S. Fearing, and H.~G. Tanner,
  ``Planning with the star (s),'' in \emph{IEEE/RSJ International Conference on
  Intelligent Robots and Systems (IROS)}, 2014, pp. 3033--3038.

\bibitem{erez2012trajectory}
T.~Erez and E.~Todorov, ``Trajectory optimization for domains with contacts
  using inverse dynamics,'' in \emph{IEEE/RSJ International Conference on
  Intelligent Robots and Systems (IROS)}, 2012, pp. 4914--4919.

\bibitem{posa2014direct}
M.~Posa, C.~Cantu, and R.~Tedrake, ``A direct method for trajectory
  optimization of rigid bodies through contact,'' \emph{The International
  Journal of Robotics Research}, vol.~33, no.~1, pp. 69--81, 2014.

\bibitem{tomic2017external}
T.~Tomi{\'c}, C.~Ott, and S.~Haddadin, ``External wrench estimation, collision
  detection, and reflex reaction for flying robots,'' \emph{IEEE Transactions
  on Robotics}, vol.~33, no.~6, pp. 1467--1482, 2017.

\bibitem{mulgaonkar2017robust}
Y.~Mulgaonkar, A.~Makineni, L.~Guerrero-Bonilla, and V.~Kumar, ``Robust aerial
  robot swarms without collision avoidance,'' \emph{IEEE Robotics and
  Automation Letters}, vol.~3, no.~1, pp. 596--603, 2017.

\bibitem{mulgaonkar2020tiercel}
Y.~Mulgaonkar, W.~Liu, D.~Thakur, K.~Daniilidis, C.~J. Taylor, and V.~Kumar,
  ``The tiercel: A novel autonomous micro aerial vehicle that can map the
  environment by flying into obstacles,'' in \emph{IEEE International
  Conference on Robotics and Automation (ICRA)}, 2020, pp. 7448--7454.

\bibitem{mintchev2017insect}
S.~Mintchev, S.~de~Rivaz, and D.~Floreano, ``Insect-inspired mechanical
  resilience for multicopters,'' \emph{IEEE Robotics and Automation Letters},
  vol.~2, no.~3, pp. 1248--1255, 2017.

\bibitem{klaptocz2013euler}
A.~Klaptocz, A.~Briod, L.~Daler, J.-C. Zufferey, and D.~Floreano, ``Euler
  spring collision protection for flying robots,'' in \emph{IEEE/RSJ
  International Conference on Intelligent Robots and Systems (IROS)}, 2013, pp.
  1886--1892.

\bibitem{naldi2014robust}
R.~Naldi, A.~Torre, and L.~Marconi, ``Robust control of a miniature ducted-fan
  aerial robot for blind navigation in unknown populated environments,''
  \emph{IEEE Transactions on Control Systems Technology}, vol.~23, no.~1, pp.
  64--79, 2014.

\bibitem{briod2014collision}
A.~Briod, P.~Kornatowski, J.-C. Zufferey, and D.~Floreano, ``A
  collision-resilient flying robot,'' \emph{Journal of Field Robotics},
  vol.~31, no.~4, pp. 496--509, 2014.

\bibitem{sareh2018rotorigami}
P.~Sareh, P.~Chermprayong, M.~Emmanuelli, H.~Nadeem, and M.~Kovac,
  ``Rotorigami: A rotary origami protective system for robotic rotorcraft,''
  \emph{Science Robotics}, vol.~3, no.~22, p. eaah5228, 2018.

\bibitem{shu2019quadrotor}
J.~Shu and P.~Chirarattananon, ``A quadrotor with an origami-inspired
  protective mechanism,'' \emph{IEEE Robotics and Automation Letters}, vol.~4,
  no.~4, pp. 3820--3827, 2019.

\bibitem{battiston2019attitude}
A.~Battiston, I.~Sharf, and M.~Nahon, ``Attitude estimation for collision
  recovery of a quadcopter unmanned aerial vehicle,'' \emph{The International
  Journal of Robotics Research}, vol.~38, no. 10-11, pp. 1286--1306, 2019.

\bibitem{dicker2017quadrotor}
G.~Dicker, F.~Chui, and I.~Sharf, ``Quadrotor collision characterization and
  recovery control,'' in \emph{IEEE International Conference on Robotics and
  Automation (ICRA)}, 2017, pp. 5830--5836.

\bibitem{briod2013contact}
A.~Briod, P.~Kornatowski, A.~Klaptocz, A.~Garnier, M.~Pagnamenta, J.-C.
  Zufferey, and D.~Floreano, ``Contact-based navigation for an autonomous
  flying robot,'' in \emph{IEEE/RSJ International Conference on Intelligent
  Robots and Systems (IROS)}, 2013, pp. 3987--3992.

\bibitem{dicker2018recovery}
G.~Dicker, I.~Sharf, and P.~Rustagi, ``Recovery control for quadrotor uav
  colliding with a pole,'' in \emph{IEEE/RSJ International Conference on
  Intelligent Robots and Systems (IROS)}, 2018, pp. 6247--6254.

\bibitem{mellinger2011minimum}
D.~Mellinger and V.~Kumar, ``Minimum snap trajectory generation and control for
  quadrotors,'' in \emph{IEEE International Conference on Robotics and
  Automation (ICRA)}, 2011, pp. 2520--2525.

\bibitem{richter2016polynomial}
C.~Richter, A.~Bry, and N.~Roy, ``Polynomial trajectory planning for aggressive
  quadrotor flight in dense indoor environments,'' in \emph{Robotics Research},
  2016, pp. 649--666.

\end{thebibliography}

\end{document}